\documentclass[letterpaper]{article} 
\usepackage{aaai23}  
\usepackage{times}  
\usepackage{helvet}  
\usepackage{courier}  
\usepackage[hyphens]{url}  
\usepackage{graphicx} 
\usepackage{natbib}  
\usepackage{caption} 
\usepackage{amsmath}
\usepackage{xspace}
\usepackage{multirow}
\usepackage{amsfonts} 
\usepackage[ruled]{algorithm2e}
\usepackage{algpseudocode}  
\usepackage{booktabs}
\usepackage{threeparttable}
\usepackage{arydshln}

\usepackage{color}

\usepackage{amsmath, bm}
\newcommand{\method}{\textsc{IRENE}\xspace}

\usepackage{newfloat}
\usepackage{listings}
\DeclareCaptionStyle{ruled}{labelfont=normalfont,labelsep=colon,strut=off} 
\title{Towards Inference Efficient Deep Ensemble Learning}
\author{
    Ziyue Li\footnote{The work was conducted during the internship of Ziyue Li at Microsoft Research Asia. Correspondence to Kan Ren.},
    Kan Ren,
    Yifan Yang,\\
    Xinyang Jiang,
    Yuqing Yang,
    Dongsheng Li
}
\affiliations{
    Microsoft Research\\

    litzy0619owned@gmail.com,
    kan.ren@microsoft.com
}

\usepackage{bibentry}

\begin{document}

\maketitle

\begin{abstract}
Ensemble methods can deliver surprising performance gains but also bring significantly higher computational costs, e.g., can be up to 2048X in large-scale ensemble tasks. 
However, we found that the majority of computations in ensemble methods are redundant. 
For instance, over 77\% of samples in CIFAR-100 dataset can be correctly classified with only a single ResNet-18 model, which indicates that only around 23\% of the samples need an ensemble of extra models. 
To this end, we propose an inference efficient ensemble learning method, to simultaneously optimize for effectiveness and efficiency in ensemble learning. 
More specifically, we regard ensemble of models as a sequential inference process and learn the optimal halting event for inference on a specific sample.
At each timestep of the inference process, a common selector judges if the current ensemble has reached ensemble effectiveness and halt further inference, otherwise filters this challenging sample for the subsequent models to conduct more powerful ensemble.
Both the base models and common selector are jointly optimized to dynamically adjust ensemble inference for different samples with various hardness, through the novel optimization goals including sequential ensemble boosting and computation saving.
The experiments with different backbones on real-world datasets illustrate our method can bring up to 56\% inference cost reduction while maintaining comparable performance to full ensemble, achieving significantly better ensemble utility than other baselines.
Code and supplemental materials are available at \url{https://seqml.github.io/irene}.
\end{abstract}

\section{Introduction}

Recent years have witnessed the great success of deep ensemble learning methods, being applied in practical machine learning applications such as image classification \cite{lee2015m,zhang2020diversified}, machine translation \cite{shazeer2017outrageously,wen2020batchensemble} and reinforcement learning \cite{yang2022towards}.
The general idea of ensemble method is to conduct the prediction upon aggregating a series of prediction outcomes from several various base models.
The benefits of ensemble method mainly lie in two aspects: improved generalization \cite{zhou2002ensembling} and specialization on different samples \cite{abbasi2020toward,gontijolopes2021representation}.

\begin{figure}[t]
\centering
\includegraphics[width=0.45\textwidth]{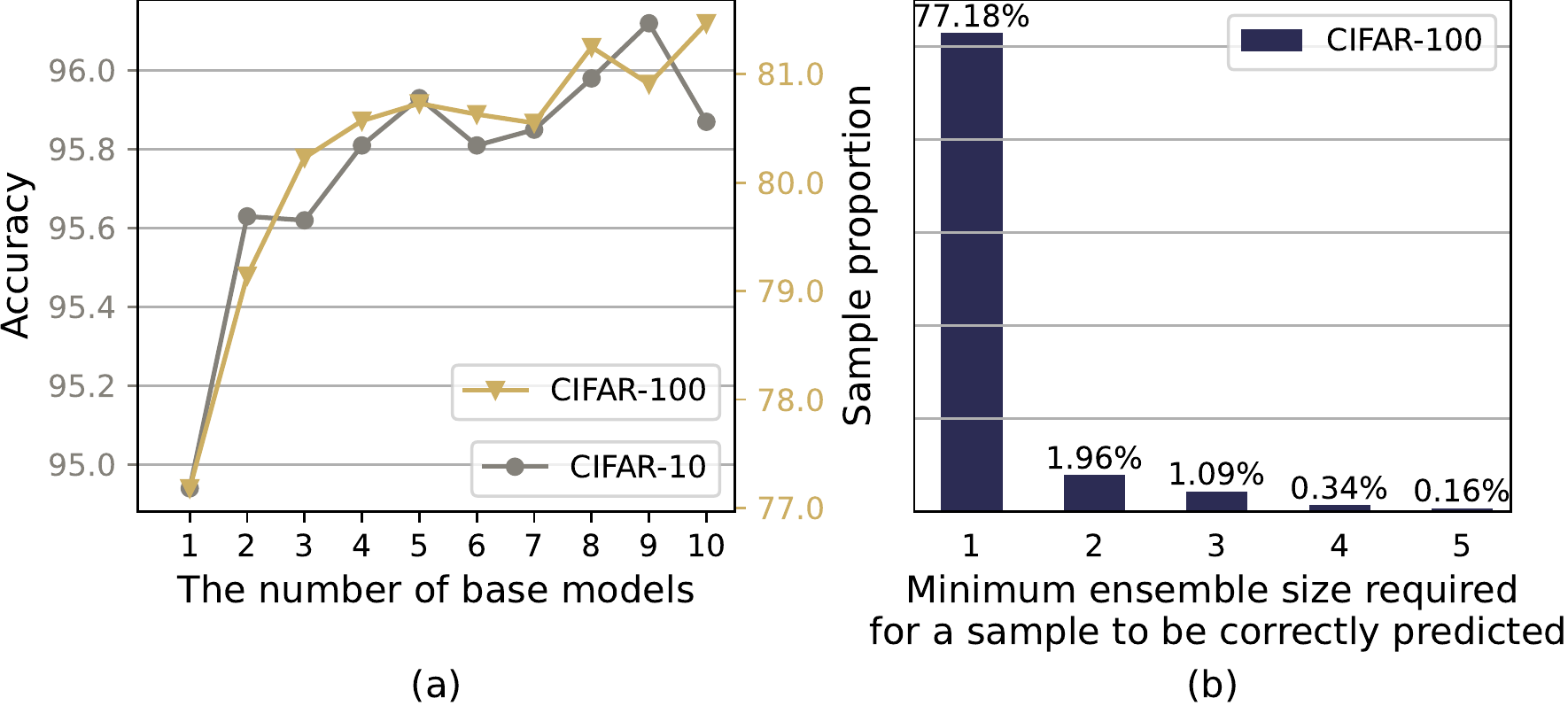} 
\caption{
(a) The performance of the average ensemble method~\cite{lakshminarayanan2016simple} on CIFAR-10/100 with the different number of base models.
(b) The minimum size of the average ensemble for  correctly predicting samples in the CIFAR-100 dataset.
}
\label{fig: k_comp}
\end{figure}

While delivering surprising performance gains, ensembles are typically much more computationally expensive compared to single model inference.
The number of the incorporated models can be up to 2048 in large-scale ensemble \cite{shazeer2017outrageously}.
However, expanding the capacity of the model pool cannot bring equal benefits.
In Figure~\ref{fig: k_comp}, we illustrate the image classification performance w.r.t. the leveraged model number in average ensemble which has been commonly used as a simple yet effective ensemble method \cite{garipov2018loss,gontijolopes2021representation}.
As shown in Figure~\ref{fig: k_comp}(a), though incorporating more base models increases the overall performance, the marginal benefit is rapidly decreasing when the number of base models is larger than 2.
Moreover, we further investigate the minimum model number required in ensemble to conduct correct prediction on CIFAR-100 \cite{krizhevsky2009learning} and Figure~\ref{fig: k_comp}(b) illustrates only a few samples need more than one model to achieve correctness.
This is interesting while reasonable since the capability of different models may overlap with each other especially when the model pool becomes larger.
We argue that it is of cost and inefficient considering the limited performance gain and the largely increased inference consumption.
Therefore, it is crucial to address the \textit{trade-off between ensemble performance and inference efficiency for different samples}, which has not received sufficient attention in previous ensemble works.

There are some works aiming to reduce computational costs of \textit{single} model inference.
One stream is pruning the model architecture \cite{han2015deep,frankle2018lottery} or quantization the network parameter \cite{liang2021pruning,jacob2018quantization}.
The other stream controls the model inference from a dynamic view \cite{han2021dynamic} by adjusting either the model depth \cite{teerapittayanon2016branchynet,huang2017multi,figurnov2017spatially} or model width \cite{shen2020fractional}.
However, these works are mainly for individual model inference and orthogonal to efficient ensemble learning.
The base models in ensemble method are individually trained and each of the base models can play an individual role of prediction, upon which ensemble has been conducted.
It is non-trivial to directly transfer the techniques of individual model optimization to ensemble structure.
The solutions to efficient ensemble learning is required.

However, efficient inferences have rarely been discussed in ensemble learning community.
One reason comes from common knowledge that more models derive higher performance gains \cite{lee2015m,shazeer2017outrageously}.
However, as we found in Figure~\ref{fig: k_comp}, the marginal benefit is lower when conducting ensemble on more models.
Some works leverage heuristic selection from the model pool, such as top-$K$ gating \cite{shazeer2017outrageously}, or stop inference based on the obtained prediction confidence in cascading ensemble \cite{wang2021wisdom}.
These studies have neither taken inference efficiency into the ensemble learning process, nor explicitly optimized for inference cost reduction. 
Thus, the adopted heuristic mechanism may derive sub-optimal solutions.

In this paper, we propose an InfeRence EfficieNt Ensemble (\method) learning approach that systematically combines ensemble learning procedure and inference cost saving as an organic whole.
Specifically, we regard ensemble learning as a sequential process which allows to simultaneously optimize training each base model and learning to select the appropriate ensemble models.
At each timestep of the sequential process, we maintain a common selector to judge if the current ensemble has reached ensemble effectiveness and halt further inference, otherwise filters this challenging sample for the subsequent models to conduct more powerful ensemble.
As for optimization, the leveraged base models and the selector are jointly optimized to (i) encourage the base models to specialize more on the challenging samples left from the predecessor(s) and (ii) keep the ensemble more effective when leveraging newly subsequent model(s) while (iii) reducing the overall ensemble costs.
Through this way, 
the extensive experiments demonstrate that our proposed method can 
reduce the average ensemble inference cost by up to 56\% while maintaining comparable performance to that on the full ensemble model pool,
outperforming the existing heuristic efficient computation methods.

\section{Related Work}

\subsection{Ensemble Learning}
\label{sec: rw_ensemble_learning}

Existing ensemble works mainly strive to improve ensemble performance, increasing model diversity to boost generalization ability.
Some methods train models in parallel and encourages the divergence of base models via incorporating additional objectives~\cite{zhou2018diverse,zhang2020diversified}.
While, others propose sequential optimization such as boosting \cite{freund1995boosting,chen2016xgboost,ke2017lightgbm}, snapshot ensemble \cite{huang2017snapshot}, and fast geometric ensembling~\cite{garipov2018loss}, to encourage model diversity by optimizing the current model on top of the existing model(s).
In general, more models yield better ensemble gains~\cite{malinin2019ensemble}. Thus, previous approaches tend to disregard the computational costs associated with ensemble ~\cite{lakshminarayanan2016simple}. 
However, the increase in benefits may be less effective compared to increased costs, as we demonstrate in this paper.

\subsection{Inference Cost Reduction}

There have been works on reducing the computational costs of a \textit{single} model.
They can be grouped into two main streams: static and dynamic solutions.
Within the former stream, numerous efforts are made to design compact network structures~\cite{sandler2018mobilenetv2,zhang2018shufflenet}, and to introduce sparsity~\cite{han2015learning}, pruning~\cite{sanh2020movement}, quantizaiton~\cite{esser2019learned} and model distillation~\cite{wang2018adversarial}, into existing networks.
Static methods treat samples of different hardness equally.
In contrast, the latter develops methods that adaptively activate some components of networks ~\cite{bengio2015conditional}, such as dynamic depth~\cite{hu2020triple,zhou2020bert}, dynamic width~\cite{bengio2013estimating} and dynamic routing~\cite{liu2018dynamic}, based on the given input.
These efficient inference methods are specifically designed for single-model architectures and are orthogonal to efficient ensemble computation that involves multiple independent models.

In ensemble learning, a few recent works consider the computational savings.
For example, several works employ heuristic criteria (e.g., confidence threshold) to stop inference~\cite{wang2021wisdom,inoue2019adaptive}.
Besides, ~\citet{shazeer2017outrageously} proposes a top-$K$ gating to determine sparse combinations of models.
These heuristics do not explicitly address efficiency as part of the learning objective, and thus may yield sub-optimal solutions.

\section{Methodology}

In this section, we first revisit the general ensemble learning framework and devise the overall optimization goal.
Then, we formulate ensemble learning as a sequential process for adaptively selecting sufficient models as an ensemble for the given sample; we define the inference efficient ensemble learning as an optimal halting problem and propose a novel framework to model this process.
Finally, we
detail the inference and optimization of our proposed method.

\subsection{Preliminaries}

Ensemble learning involves training multiple base models that perform diversely and aggregating their predictions. 
This can lead to significant performance improvements, but also a corresponding increase in computational cost through the inclusion of various models.

Therefore, considering the computational cost, the general objective of ensemble is to maximize the performance and minimize the inference cost in expectation as
\begin{equation}\label{eq:simple_objective}
\small
\begin{aligned}
\max_{\left\{\theta_t\right\}^{T}_{t=1}}{\mathbb{E}}_{(x,y) \sim D}
\left[  V(y,\left\{\hat{y}_t\right\}^{T}_{t=1})  - \alpha C(\left\{\theta_t\right\}^{T}_{t=1}|x) \right],
\end{aligned}
\end{equation}
where $V$ measures ensemble performance, $C$ measures the inference cost 
given a specific sample, and $\alpha$ weighs these two objectives.
Here the prediction of $t$-th model $f_{\theta_t}(\cdot)$ models
${\rm Pr}(y|x)$ and $\hat{y}_t = f_{\theta_t}(x)$ where $\theta_t$ is the corresponding model parameter.
$D$ is the given dataset.

\noindent\textbf{Realization.}
Here we discuss some specific formulations to realize the performance and cost metrics.
To measure ensemble performance, generally task-specific loss $\mathcal{L}$ can be adopted, such as cross-entropy loss for classification and mean squared error for regression~\cite{zhang2020diversified}.
While various metrics, e.g., latency or FLOPs~\cite{wang2021wisdom}, can be utilized for measuring inference cost.

However, most works only focus on optimizing ensemble performance, by striving to obtain diverse base models for improving the generalization of ensembles ~\cite{lee2015m}.

As for the inference cost, the existing ensemble works only perform heuristic model selection to save computational effort.
For example, some works either choose a fixed number of top-ranked predictions for all the samples~\citep{shazeer2017outrageously}, or select a subset of models whose ensemble has prediction confidence above a predefined threshold~\citep{wang2021wisdom}.
These heuristic solutions may yield sub-optimal solutions in terms of performance and efficiency.
Instead, we turn to a joint consideration of ensemble performance and efficiency, modeling the optimization of efficiency as part of ensemble learning.

\subsection{Sequential Ensemble Framework}
In this work we attempt to model ensemble inference as a sequential process in which a learning-based model selection yields a sample-specific efficient ensemble over a subset of models.
Specifically, during the process, model inference is sequential and selectively halts at one timestep.
In this section, we first discuss two paradigms of ensemble inference and their pros and cons, then we present the model selection problem, introduce our design for sequential model selection, and describe the entire inference process.

\subsubsection{Sequential inference v.s. parallel inference}

\begin{figure}[ht]
\centering
\includegraphics[width=0.47\textwidth]{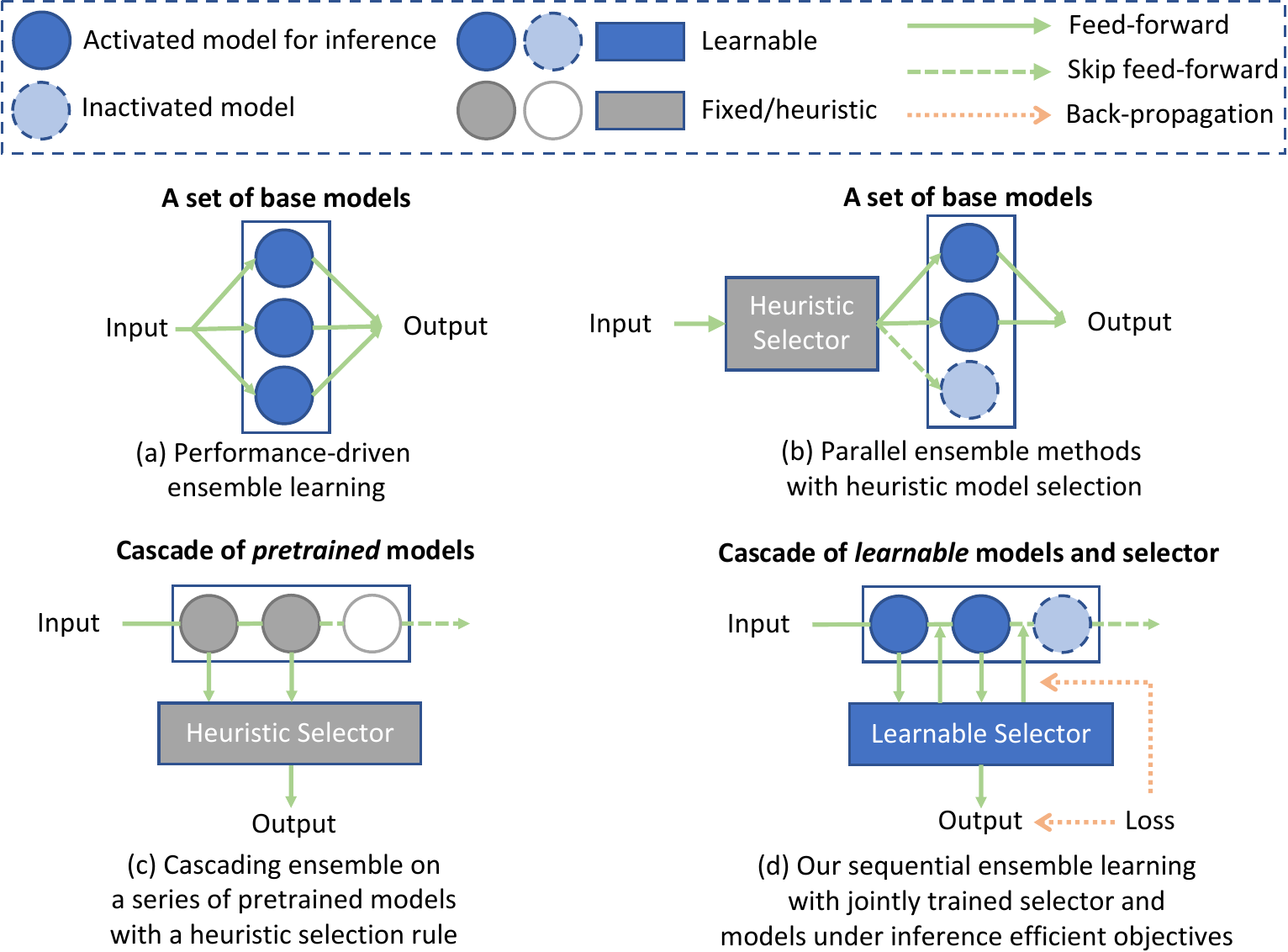} 
\caption{
Comparison of (a) the general ensemble, (b) the approach using heuristic model selection under the parallel paradigm and (c) under the sequential paradigm, and (d) our approach.
Our approach differs from the existing methods in two distinct ways: 
(i) we build a learnable selector to explicitly optimize the inference cost, and 
(ii) model training and selection learning occur as an organic whole.
}
\label{fig: paradigm_comp}
\end{figure}

The inference paradigms of ensemble methods with computation saving can be divided into two types: parallel and sequential, as shown in Figure~\ref{fig: paradigm_comp}(b) and (c).
In the parallel paradigm, there is a module that determines which models are activated for inference \textit{prior to} model inference~\cite{shazeer2017outrageously}.
Therefore, the framework does not utilize actual outcomes or the inference situation of base models to decide (which models would be selected).
In contrast, in the sequential paradigm, outcomes can be taken at any step throughout the inference process~\cite{wang2021wisdom} to determine how to balance the effectiveness and efficiency of the ensemble.
Thus, we focus on the sequential paradigm, which allows using model-relevant information to dictate methods for balancing ensemble effectiveness and efficiency, in this paper.

\begin{figure}[t]
\centering
\includegraphics[width=0.45\textwidth]{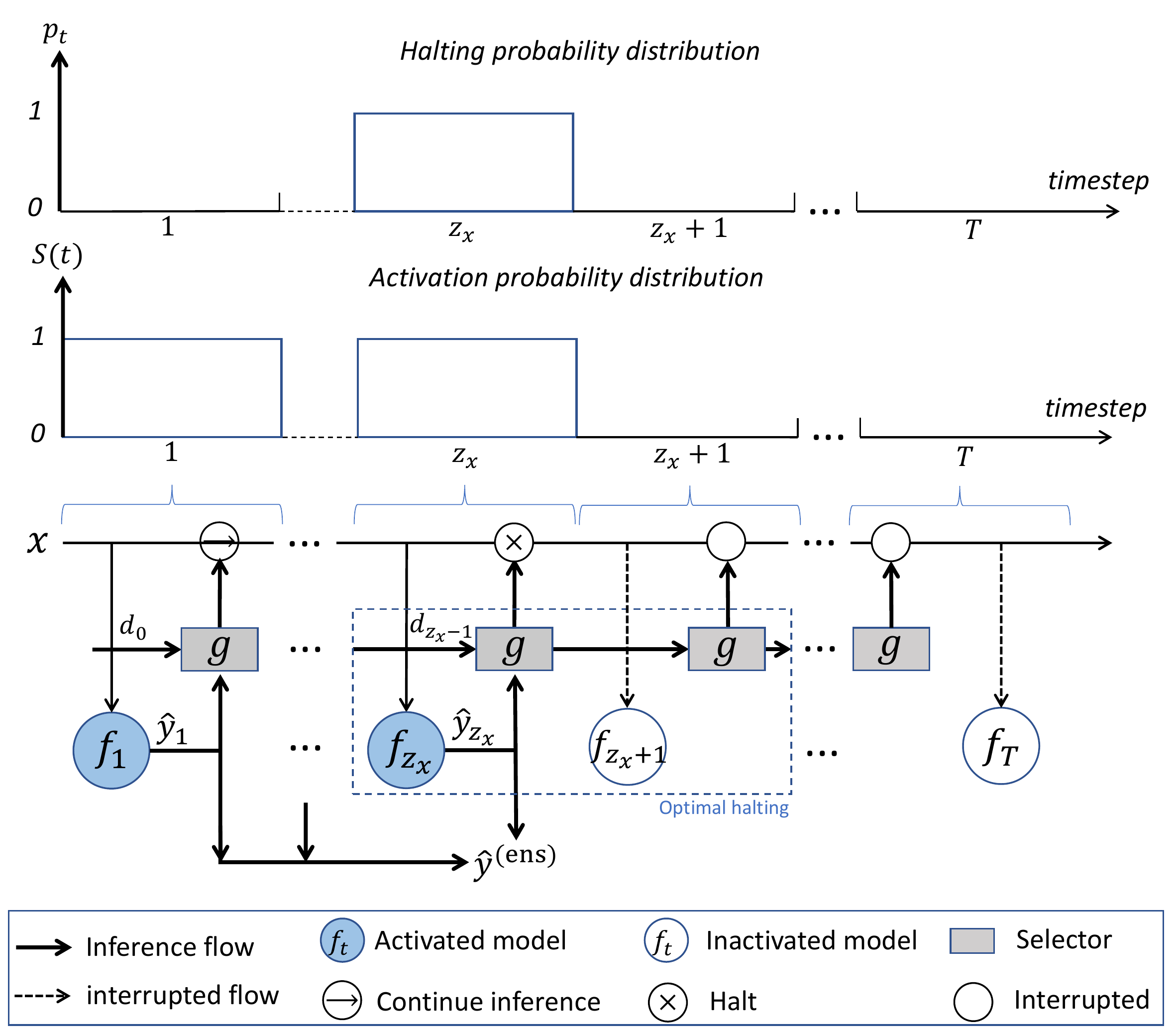} 
\caption{
Illustration of our sequential inference and optimal halting mechanism.
}
\label{fig:framework}
\end{figure}

As illustrated in Figure~\ref{fig:framework}, in our framework, all the base models are kept ordered for training (ensemble learning) and inference (ensemble inference).
In the inference process for a given sample, each base model will be activated for inference until the optimal halting event occurs, which has been decided by the jointly trained selector.
And the predictions of all activated models will be aggregated as output.

\subsubsection{Optimal halting}

Here we define the optimal halting event and describe our inference halting mechanism.

We first illustrate the notations and the settings of our sequential inference and optimal halting mechanism.
Specifically, we address the problem of selecting an appropriate subset of models for a given sample. 
Suppose that there are $T$ models cascaded together; and the inference of the base model inference is performed sequentially once at each timestep.
At each timestep $t$, the $t$-th model will be executed and one selector shared by all the timesteps will decide to halt at the current timestep or continue inference for further ensemble.
Given one specific sample, halting at a certain timestep implies that, the predictions of the models \textit{at this step and before} will be aggregated as the final ensemble output, and leveraging more following base models will not help improve the prediction performance while only increasing the inference cost.

Formally, given such an ordered sequence of models, we define the optimal halting timestep as 
$z$ at which it is optimal to halt for some criterion, i.e., the overall optimization objective in Eq.~(\ref{eq:simple_objective}) has been achieved.
We define the probability distribution $p_t$ on $\left\{1, \dots, T\right\}$  to represent the probability that halting at the $t$-th step is optimal.

With probability distribution $p_t$, we can define probability that the optimal halting step is \textit{at or after} the $t$-th step as
\begin{equation}
	\label{eq:S(t)}
	\resizebox{0.83\columnwidth}{!}{
\begin{math}
    \begin{aligned}
    S(t) = {\rm Pr}(z\geq t) = 1 - {\rm Pr}(z < t) = 1 - \sum\nolimits^{t-1}_{i=1} p_i,
    \end{aligned}
    \end{math}}
\end{equation}
which is calculated from the cumulative probability of the optimal halting step.
Note that, according to our setting, all models \textit{before and at} the optimal halting step are activated for inference.
Thus, the $t$-th model is activated for inference when the optimal halting step is not within the first $(t-1)$ steps, i.e., $S(t)==1$.
From the definition, we can say that $S(t)$ represents the probability of activating all the models ordered \textit{before and at} $t$.

We define the conditional probability $h_t$ of halting at the $t$-th step given that no halting has occurred before as
\begin{equation}\label{eq:Pr(z=t|z>t-1)}
\resizebox{0.905\columnwidth}{!}{
\begin{math}\begin{aligned}
  h_t = {\rm Pr}(z=t|z>t-1) = & \frac{{\rm Pr}(z=t)}{{\rm Pr}(z>t-1)}
    = \frac{S(t)-S(t+1)}{S(t)}.
\end{aligned}\end{math}}
\end{equation}

To determine whether to halt at each timestep $t$, our goal is to predict the probability $p_t$ of the optimal halting event.
And the probability will be further utilized for 
efficiency-aware ensemble inference.

We can further derive the calculation of the probability functions $S(t)$ based on the conditional probability $h_t$ at each timestep.
The probability that, the optimal halting step is larger than or equal to the current timestep $t$, can be calculated following a probability chain as
\begin{equation}\label{eq:S(t)_}
\resizebox{0.8\columnwidth}{!}{
\begin{math}\begin{aligned}
  S(t) = & {\rm Pr}(z\neq t-1, \dots, z\neq 1) \\
    = & {\rm Pr}(z\neq t-1|z>t-1)\cdots{\rm Pr}(z\neq 1|z>0) \\
    = & \prod\nolimits^{t-1}_{i=1} (1-h_i).
\end{aligned}\end{math}}
\end{equation}
Based on Eq.~\eqref{eq:Pr(z=t|z>t-1)} and \eqref{eq:S(t)_}, we can derive the probability $p_t$ as
\begin{equation}\label{eq:p_t}
\begin{aligned}
  p_t = h_t \prod\nolimits^{t-1}_{i=1}(1-h_i).
\end{aligned}
\end{equation}
Therefore, we can predict whether to halt at a step based on the predicted conditional probability $h_t$ prior to and at the timestep to be estimated.

\subsubsection{Sequential halting decision}

Based on our derived probabilistic formulation of the optimal halting
problem, we propose the selector network that leverages sample and model-related information to predict the conditional probability $h_t$ for the calculation of halting probability $p_t$ at each timestep.

The selector network $g_{\phi}$ should be of the form $h_t, d_t = g_{\phi}(e(x),d_{t-1})$, to generate state-specific information based on the sample-related information encoded by $e(\cdot)$ and the output $d_{t-1}$ of the last timestep.
Although various types of neural networks can be applied for $g$, we choose the recurrent neural network which is commonly used for modeling conditional probabilities over time~\cite{schuster1997bidirectional}.
Besides, we take the output of the $t$-th model $f_{\theta_t}$, which provides both information related to the sample and the $t$-th model that has participated in ensemble inference, as $e(x)$.
The implementation details refer to Appendix A.1.

\subsubsection{Ensemble inference}

Here we detail the overall inference process.
Given a specific sample $x$, the $t$-th model and selector infer in turn at each timestep $t$.
Specifically, the base model will be activated to produce the individual prediction $\hat{y}_t$, and the selector will compute the optimal halting probability $p_t$ to determine if inference of the given sample should be continued ($p_t==0$) to subsequent base models $\{f\}_{t+1}^T$, or stop the inference ($p_t==1$).
Based on previous definitions, we can derive the halting step, on the basis of which ensemble prediction and ensemble efficiency are calculated.
The optimal halting step $z_{x}$ for the sample $x$ satisfies that
\begin{equation}\label{eq:z_x}
\begin{aligned}
  z_{x} = \mathop{\arg\max}\limits_{t} p_t = \mathop{\arg\max}\limits_{t} (h_t \prod\nolimits^{t-1}_{i=1} (1-h_i)).
\end{aligned}
\end{equation}
To sample a unique $z_{x}$ from $p_t$, we first sample each $h_t$ using a differentiable sample from the Gumbel-Softmax distribution~\cite{jang2016categorical}.
Specifically, after $h_t$ is sampled and binarized using a trainable binary mask in~\cite{mallya2018piggyback}, only one timestep $z_x$ has the maximum probability value ($p_{z_x}= 1.0$) \textcolor{black}{when $h_{z_x}= 1$, and $h_t=0$ for any $t<z_x$}. \textcolor{black}{That is,} $p_t$ of any timestep $t$ after $z_x$, being the product of \textcolor{black}{$(1-h_{z_x})$} with value zero and any number, will be zero in Eq.~\eqref{eq:p_t}.

As halting at the $z_x$-th timestep, the ensemble prediction 
can be calculated as an aggregation from the previous base model outputs as
\begin{equation}\label{eq:y_ens}
\begin{aligned}
  \hat{y}^{\text{(ens)}} = \text{ENS}\left( \{ \hat{y}_t \}_{t=1}^{z}\left\vert _{z = z_{x}} \right. \right) =\frac{1}{z_{x}} \sum\nolimits^{z_{x}}_{t=1} \hat{y}_t .
\end{aligned}
\end{equation}
Here we simply utilize average ensemble \cite{huang2017snapshot,garipov2018loss,wang2021wisdom} 
as an example of ensemble aggregation ENS.
Note that in our implementation, all base models use the same backbone, thus, the ensemble efficiency can be directly measured as the number of steps taken $z=z_x$.

\subsection{Ensemble Learning}

Note that, we have \textit{no} ground truth of the optimal halting step $z$ for each sample, otherwise we could directly model that through maximum likelihood estimation.
Therefore, we propose to optimize our sequential ensemble framework from the perspective of maximizing ensemble performance as well as ensemble efficiency.
In the following, we will first introduce the optimization of base model performance, then the ensemble performance and efficiency, and then introduce inference efficient ensemble optimization by jointly optimizing base models and the selector.

\subsubsection{Optimization of base model performance}

As the cornerstone of the entire framework, the performance of the base models is crucial.
We train each base model by minimizing the task-specific loss $\mathcal{L}^{\text{(base)}}_t$ over all training samples as
\begin{equation}\label{eq:base_loss}
\begin{aligned}
  \min_{\theta_t} &~ {\mathbb{E}}_{(x,y)  \sim D} \mathcal{L} (y, 
  \hat{y}_t
  ).
\end{aligned}
\end{equation}
\textcolor{black}{We will describe its joint optimization with the selector later.}

\subsubsection{Optimization of ensemble performance}

Ensemble performance is a critical goal.
In particular, we optimize this goal at each timestep to ensure that the halting strategy is optimized
toward the final objective.

At each $t$-th step, the cascade yields its predictions for the current stage, by averaging the predictions of all the activated models for the given sample as
\begin{equation}
	\label{eq:early_ensemble}
	\resizebox{0.7\columnwidth}{!}{
\begin{math}
    \begin{aligned}
    \hat{y}^{(\text{ens})}_t =
    \left(\sum\nolimits^{t}_{i=1} (S(i)\hat{y}_i)\right)/\left(\sum\nolimits^{t}_{i=1} S(i)\right),
    \end{aligned}
    \end{math}}
\end{equation}
where $S(i)$ defined in Eq.~\eqref{eq:S(t)} indicates the probability of $\hat{y}_i$ will be aggregated into the ensemble.
Different from Eq.~\eqref{eq:y_ens} which computes the final ensemble prediction when the halting step $z_x$ is reached, Eq.~\eqref{eq:early_ensemble} computes the (temporary) ensemble predictions obtained at any step, regardless of when the halting step occurs.

We optimize the task-specific loss of ensemble prediction $\mathcal{L}^{\text{(ens)}}_t$ at each timestep as
\begin{equation}\label{eq:ensemble_loss}
\begin{aligned}
  \min_{\phi} &~ {\mathbb{E}}_{(x,y)  \sim D}\left[ \mathcal{L}(y, \hat{y}^{(\text{ens})}_t) \right].
\end{aligned}
\end{equation}
Note that, this objective is only used to optimize the selector parameter $\phi$ because it may hinder model training, as proven in \citet{allen2020towards}: optimizing base models by ensemble performance largely degrades performance.

\subsubsection{Optimization of ensemble inference efficiency}

The selector should be aware of the inference cost imposed by its own strategy to discourage trading full usage for higher performance.
In this work, we directly use the number of steps taken before halting (i.e., the ensemble size for a sample) as a measure of inference cost $\mathcal{L}^{\text{(cost)}}_t$ and optimize as
\begin{equation}\label{eq:cost_loss}
\resizebox{0.88\columnwidth}{!}{
\begin{math}
\begin{aligned}
  \min_{\phi} &~ {\mathbb{E}}_{(x,y)  \sim D, \atop
  \left\{h_i \sim g_{\phi}(\cdot,i)\right\}^{t-1}_{i=1},}
  \left[ \sum\nolimits^{T}_{t=1}(t\cdot(\underbrace{h_t \prod\nolimits^{t-1}_{i=1} (1-h_i))}_{p_t})\right].
\end{aligned}
\end{math}
}
\end{equation}
Since all the models in our framework use the same backbone, as a common practice in ensemble learning, the number of activated models for ensemble is a direct measurement of the inference cost.
If the base models use different network backbones, the measurement could be other metrics such as the activated model parameters or inference FLOPs, which does not influence our proposed approach.

\subsubsection{Joint optimization for effective and efficient ensemble}

During the sequential inference process in our ensemble method, the selector relying on the inference situation tries to decide whether or not to halt (stop inference for ensemble) at the current timestep.
Thus, the whole system (both selector and the base models) should balance the performance gain when incorporating more base models with the increased inference cost.

We expect that adding a model to the ensemble leads to an improvement in performance, which means that the newly added model is more specialized on these samples.
To this end, one way is, for the selector, to adjust its selection of models and the other way is to encourage each base model to be more focusing on the samples to which they are assigned (by the sequential inference process) accordingly.

We propose an objective $\mathcal{L}^{\text{(rank)}}_t$ to optimize from both perspectives simultaneously as
\begin{equation}\label{eq:ranking_loss}
\resizebox{0.885\columnwidth}{!}{
\begin{math}
\begin{aligned}
  \min_{\theta_t, \phi} &~ {\mathbb{E}}_{(x,y)  \sim D}
   \left[ \max (0,S(t)(\mathcal{L}(y,\hat{y}_t)-\hat{\mathcal{L}}(y,\hat{y}^{\text{(ens)}}_{t-1}) ) \right],
\end{aligned}
\end{math}}
\end{equation}
where $S(t)$ in Eq.~\eqref{eq:S(t)} denotes 
the probability that
inference does not halt at reaching the $t$-th step.
Note that, with $S(t)$ equal to 1, the right-hand element of the maximum function will be the difference between the loss $\mathcal{L}(y,\hat{y}_t)$ of the $t$-th model and the referential loss $\hat{\mathcal{L}}(y,\hat{y}^{\text{(ens)}}_{t-1})$ of the ensemble of the first $(t-1)$ models. 
$\hat{\mathcal{L}}$ plays a role as the target value for bootstrapping \textit{without} backpropagation at timestep $t$.
Otherwise when the difference is less than 0 or $S(t)=0$, the objective output is 0, as it implies the loss of the $t$-th model being less than that of the other one; or inference has been halted before the $t$-th step.

On one hand, $\mathcal{L}^{\text{(rank)}}_t$ encourages the potentially incorporated base models to obtain lower task-specific losses, i.e., better performance, on the sample that is assigned to them than \textit{the ensemble of the previous models}.
On the other hand, it incents the selector to let subsequent models get samples they are better at than \textit{the ensemble of the previous models}.

\subsubsection{Sequential training paradigm}

Aligned with the sequential inference process, the training process is also sequentially conducted.
Specifically, 
we initialize the sequentially ordered base models and optimize them step by step.
Based on our proposed optimization objectives,
we minimize the expected task-specific loss and inference cost by jointly optimizing base models and the selector at each timestep $t$ as
\begin{equation}\label{eq:total_loss}
\begin{aligned}
  \mathcal{L}^{(\text{total})}_t = \underbrace{\mathcal{L}^{\text{(base)}}_t}_{\text{model optimization}}   + \underbrace{\omega_1 \mathcal{L}^{\text{(ens)}}_t +  \omega_2 \mathcal{L}^{\text{(cost)}}_t}_{\text{selector optimization}}  + \underbrace{\omega_3 \mathcal{L}^{\text{(rank)}}_t}_{\text{joint optimization}},
\end{aligned}
\end{equation}
where $\omega_1$, $\omega_2$, and $\omega_3$ are the loss weights.
The overall training algorithm has been illustrated in Appendix B.

Together, these objectives serve to learn an inference efficient ensemble.
The optimization of a single base model for $\mathcal{L}^{\text{(base)}}_t$ is relatively independent from the other objectives, but is also the basis of our entire framework.
For the selector optimization, the two objectives applied are adversarial, as optimizing $\mathcal{L}^{\text{(ens)}}_t$ readily employs all models in exchange for advanced performance, while optimizing $\mathcal{L}^{\text{(cost)}}_t$ may decrease the ensemble performance while saving inference costs.
Though these two objectives can in essence optimize both ensemble performance and ensemble efficiency, it is difficult to expect the selector to work reasonably well, e.g., to perform further inference only when subsequent models can bring improvements.
Therefore, we further propose $\mathcal{L}^{\text{(rank)}}_t$ to jointly optimize an effective and efficient ensemble, ensuring that adding more models is beneficial and models are more focused on samples assigned to them.

\subsubsection{Discussion: Paradigm differences from prior methods}

Our ensemble paradigm varies to existing paradigms in the optimal halting strategy learned from the interaction of the selector and base models via their joint optimization, as shown in Figure~\ref{fig: paradigm_comp}.
From a paradigm perspective, we introduce a learning-based selector rather than heuristics.
In addition, our selector interacts with base models to facilitate their attention to samples assigned to them, instead of receiving model outcomes to independently determine the halting step.
From the optimization perspective, we propose novel objectives for inference efficient ensemble learning as discussed above.
Note that, by optimizing Eq.~\eqref{eq:ranking_loss}, we realize the interaction between the base models and the selector, rationalizing the strategies made by the selector and aligning the performance of the base models with the selection.

\section{Experiment}

\subsection{Experimental setup}

Here we present the details of experimental setup, including datasets, backbones used, and baselines for comparison.

\noindent
\textbf{Datasets and backbones.}
We conduct experiments on two image classification datasets, CIFAR-10 and CIFAR-100, which are the primary focus of neural ensemble methods ~\cite{zhang2020diversified,rame2021dice}.
CIFAR~\cite{krizhevsky2009learning} contains 50,000 training samples and 10,000 test samples, which are labeled as 10 and 100 classes in CIFAR-10 and CIFAR-100, respectively.
Following the previous setup of ensemble methods,
we adopt ResNet-32 and ResNet-18~\cite{he2016deep} as backbones and all the ensemble methods to be compared use three base models, i.e., $T=3$.
We also provide ablation studies on $T$.

\noindent
\textbf{Baselines.}
We compare \method with various ensemble methods, the implementation details of which are given in Appendix A.2.
Traditional ensemble methods that do not consider costs, includes \textbf{MoE}~\cite{shazeer2017outrageously}, \textbf{average ensemble},
\textbf{Snapshot ensemble}~\cite{huang2017snapshot} and \textbf{fast geometric ensembling}~\cite{garipov2018loss} (FGE).
Other than them, we also compare with \textbf{Sparse-MoE} ~\cite{shazeer2017outrageously} and  \textbf{WoC}~\cite{wang2021wisdom}, both of which adopt heuristic computation saving methods.
For Sparse-MoE, using the same setup as theirs, two models are activated for each sample in the inference.
For WoC, which uses a confidence threshold-based halting on a cascade of pretrained models, we follow them and implement it on the trained base models of average ensemble.

\noindent
\textbf{Evaluation metrics.}
We report, for each method, its
utility value (the trade-off between ensemble performance and efficiency), top-1 accuracy, and the corresponding inference cost, i.e., average number of utilized models in ensemble. Their detailed description is in Appendix C.

\subsection{Evaluation results}

\begin{table}[t]
\centering
\setlength{\belowcaptionskip}{-5pt}
\resizebox{.45\textwidth}{!}{
\begin{tabular}{lllll}
\hline
& Methods & Top-1 (\%) $\uparrow$ & Cost $\downarrow$ & Utility $\uparrow$ \\ \hline
\multirow{9}{*}{\rotatebox{90}{ResNet-32}} & Single model &    93.10  $_{\pm 0.12}$       &     1.00           &   1.00      \\ \cdashline{2-5}
& Average ensemble                                          &  \textbf{94.46} $_{\pm 0.13}$   &        3.00     &   1.00     \\ 
& Snapshot ensemble  &   93.73 $_{\pm 0.36}$      &    3.00      &   0.56 \\
& FGE  & 93.19  $_{\pm 0.11}$ &    3.00   &   0.36    \\ 
& MoE   &  93.26 $_{\pm 0.35}$  & 3.00   &   0.38  \\ 
& Sparse-MoE & 81.31 $_{\pm 2.50}$  &  2.00   &   0.00\\ 
& WoC &   94.29 $_{\pm 0.00}$   & 2.48 $_{\pm 0.09}$   &  1.05    \\ 
& \method   & \textbf{94.46} $_{\pm 0.07}$ &    \textbf{2.21} $_{\pm 0.12}$   &    \textbf{1.37}\\ \hline \hline
\multirow{8}{*}{\rotatebox{90}{ResNet-18}} & Single model &    94.94 $_{\pm 0.07}$       &     1.00           &   1.00      \\  \cdashline{2-5}
& Average ensemble                                          &  95.62 $_{\pm 0.04}$      &    3.00 & 1.00      \\ 
& Snapshot ensemble  & 95.61 $_{\pm 0.14}$  &     3.00    &   0.98   \\
& FGE  & 94.67 $_{\pm 0.06}$  &  3.00   &   0.22    \\ 
& MoE   & 94.41 $_{\pm 0.06}$   &       3.00  &   0.14 \\ 
& Sparse-MoE & 84.07  $_{\pm 1.42}$ & 2.00         &  0.00\\ 
& WoC &   95.48 $_{\pm 0.01}$ &  \textbf{1.15} $_{\pm 0.02}$       &    2.10  \\ 
& \method & \textbf{95.81} $_{\pm 0.08}$ &  1.32 $_{\pm 0.13}$       &    \textbf{3.10}\\ \hline
\end{tabular}
}
\caption{
Experiment results on CIFAR-10.
}
\label{tab: cifar10}
\end{table}

We demonstrate the effectiveness of \method on two benchmark datasets, CIFAR-10 and CIFAR-100, using two different backbones, with results shown in Table~\ref{tab: cifar10} and \ref{tab: cifar100}.
We mark the best results in bold with arrows ($\uparrow$/$\downarrow$) indicating the direction of better outcomes for the metrics.

\subsubsection{Performance comparison}
\quad

\noindent
\emph{\method achieves better trade-offs than traditional ensemble methods and heuristic cost-constrained methods.}
While comparing with traditional methods with fixed costs, we are interested in the inference cost saved by \method.
As shown in tables, the inference costs of \method  are 73.67\%, 44.00\%, 84.33\%, and 62.67\% of those in traditional ensemble methods, resp.
That is, a large fraction of the inference costs can be saved while the performance penalty turns out to be small or even negligible.
Accordingly, the utility of \method is higher than average ensemble by an average of 78.25\%.
Compared with Sparse-MoE and WoC, two methods that use heuristic computation saving, \method also scores higher in utility, indicating that it is superior to heuristic solutions.

\begin{table}[t]
\centering
\resizebox{.43\textwidth}{!}{
\begin{tabular}{lllll}
\hline
& Methods & Top-1 (\%) $\uparrow$ & Cost $\downarrow$ & Utility $\uparrow$ \\ \hline
\multirow{8}{*}{\rotatebox{90}{ResNet-32}} & Single model &    69.58 $_{\pm 0.52}$       &     1.00           &   1.00      \\ \hdashline
& Average ensemble                                          &  \textbf{74.94}  $_{\pm 0.30}$   &        3.00     &   1.00   \\ 
& Snapshot ensemble  &   74.26  $_{\pm 0.18}$      &    3.00      &  0.97   \\
& FGE  & 71.19  $_{\pm 0.27}$ &    3.00   &   0.86  \\ 
& MoE   &  70.64  $_{\pm 1.00}$  & 3.00   & 0.84  \\ 
& Sparse-MoE & 49.48 $_{\pm 1.09}$  &  2.00   & 0.00 \\ 
& WoC &   73.90 $_{\pm 0.24}$  &  \textbf{2.31} $_{\pm 0.21}$   &  1.05     \\ 
& \method                                      & 74.84 $_{\pm 0.06}$ &    2.53 $_{\pm 0.08}$   &    \textbf{1.16}\\ \hline \hline
\multirow{9}{*}{\rotatebox{90}{ResNet-18}} & Single model &    77.18 $_{\pm 0.16}$       &     1.00           &   1.00      \\ \hdashline
& Average ensemble                                          &  \textbf{80.28} $_{\pm 0.25}$      &    3.00 &  1.00    \\ 
& Snapshot ensemble  & 79.17 $_{\pm 0.19}$   &     3.00    &  0.68   \\
& FGE  & 77.84 $_{\pm 0.37}$  &  3.00   &  0.42 \\ 
& Div$^{2}$ & 79.12
& 3.00 & 0.67 \\
& MoE   & 77.49 $_{\pm 0.37}$   &       3.00  &   0.37 \\ 
& Sparse-MoE & 59.04  $_{\pm 0.82}$ & 2.00         &  0.00 \\ 
& WoC &   79.86 $_{\pm 0.01}$ &  1.93   $_{\pm 0.06}$       &   1.35 \\
& \method &  80.10 $_{\pm 0.14}$ &    \textbf{1.88} $_{\pm 0.05}$       &    \textbf{1.50}\\ \hline
\end{tabular}
}
\caption{
Experiment results on CIFAR-100.
}
\label{tab: cifar100}
\end{table}

\begin{figure}[ht]
\includegraphics[width=0.45\textwidth]{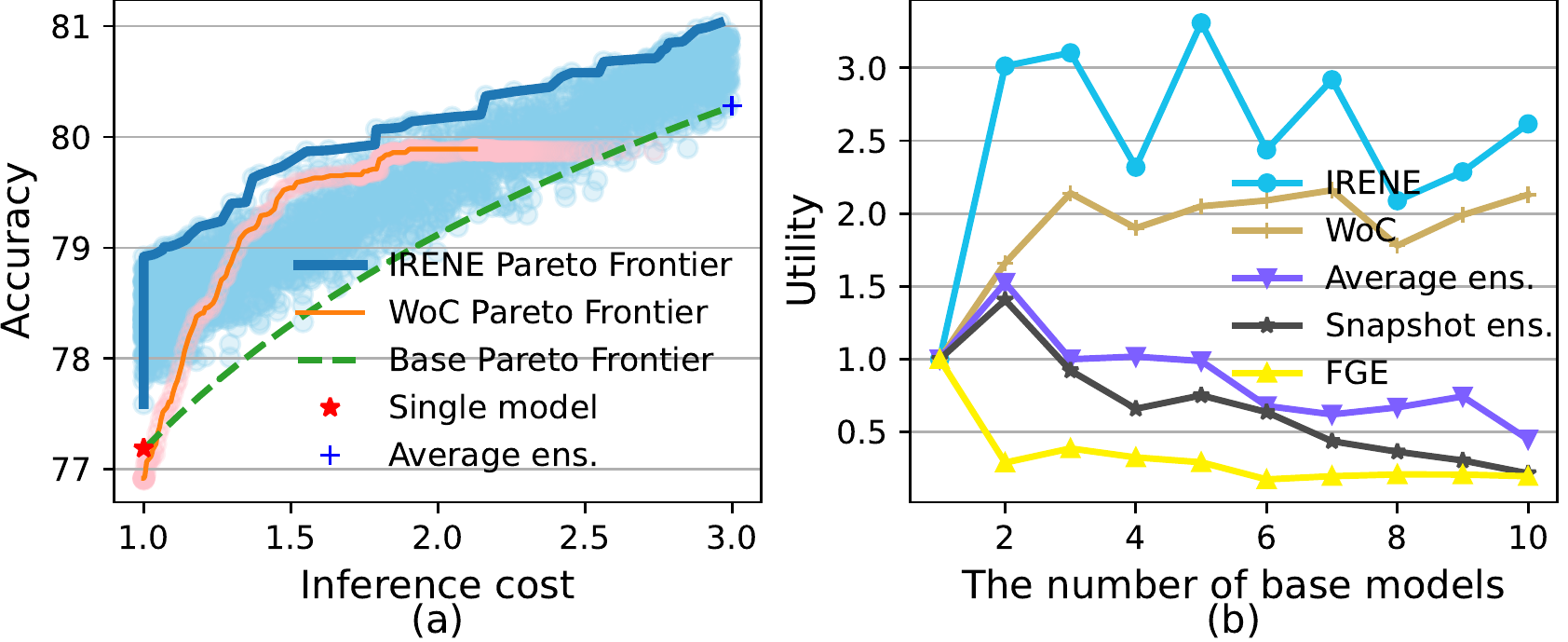} 
\caption{
(a) Pareto frontier on CIFAR-100 using ResNet-18. 
The curve of single model (average ensemble) is denoted as base Pareto frontier.
(b) Utility with varied number of ResNet-18 models on CIFAR-10. 
}
\label{fig: perato_curve}
\end{figure}

\noindent
\emph{Improvement of \method is related to dataset difficulty and neural network capability.}
As shown in Table~\ref{tab: cifar10}, the relatively advanced backbone ResNet-18 shows a surprising result over the easier CIFAR-10 dataset.
Specifically, the inference cost of \method is reduced by 56.0\% versus average ensemble, while its performance is improved by 0.19\%.
This supports that the existence of redundancy between models and \method seizes it for efficient and effective inference.
Additionally, both backbones use less inference cost in CIFAR-10 than in CIFAR-100, which is reasonable since the task is relatively more difficult in CIFAR-100 and it also illustrates that our proposed method of inference efficient ensemble learning can dynamically adjust the ensemble efficiency according to the task difficulty.

\subsubsection{Sensitivity analysis A: The Pareto frontier}

\noindent
We compare \method, WoC, and average ensemble for their performance-cost trade-off through the Pareto frontier, and their results using ResNet-18 on CIFAR-100 are shown in Figure~\ref{fig: perato_curve}(a).
From the figure, \textit{for a small increase in cost (around 1.0), \method achieves a significant improvement in accuracy}.
In addition, \method competitively obtains Pareto optimal values under all cost regimes.
Moreover, \method yields significantly better performance than traditional ensemble methods under various cost constraints, indicating \method benefits performance-cost trade-off as well as model training.
In contrast, WoC can only approach the optimum in a limited cost region and leaves a clear cutoff point in the curve, suggesting that its heuristic solution is sub-optimal since adding more inference cost does not bring promising performance gains which also illustrates that our proposed method is superior for effective and efficient ensemble learning.

\subsubsection{Sensitivity analysis B: Utility of \method with different number of base models}

\noindent
\method aiming to learn inference efficient ensemble is expected to achieve better trade-off between performance and cost thus preventing introducing worthless inference cost.
To verify this, we compare the utility of various ensemble methods when the number of base models varies, with CIFAR-10 results using ResNet-18 shown in Fig.~\ref{fig: perato_curve}(b).
The other compared methods
include average ensemble, snapshot ensemble, FGE, and WoC.

An insight drawn from the results is that, as the number of models increases, the utility of traditional ensemble methods generally first increases and then continues to decline.
This further suggests that existing ensemble methods may be trading unnecessary computation for performance gains.
\method and WoC, in contrast, can maintain decent utility despite adding more base models and \method outperforms WoC by a large margin, proving the robustness and effectiveness of \method.

\begin{figure}[t]
\centering
\includegraphics[width=0.5\textwidth]{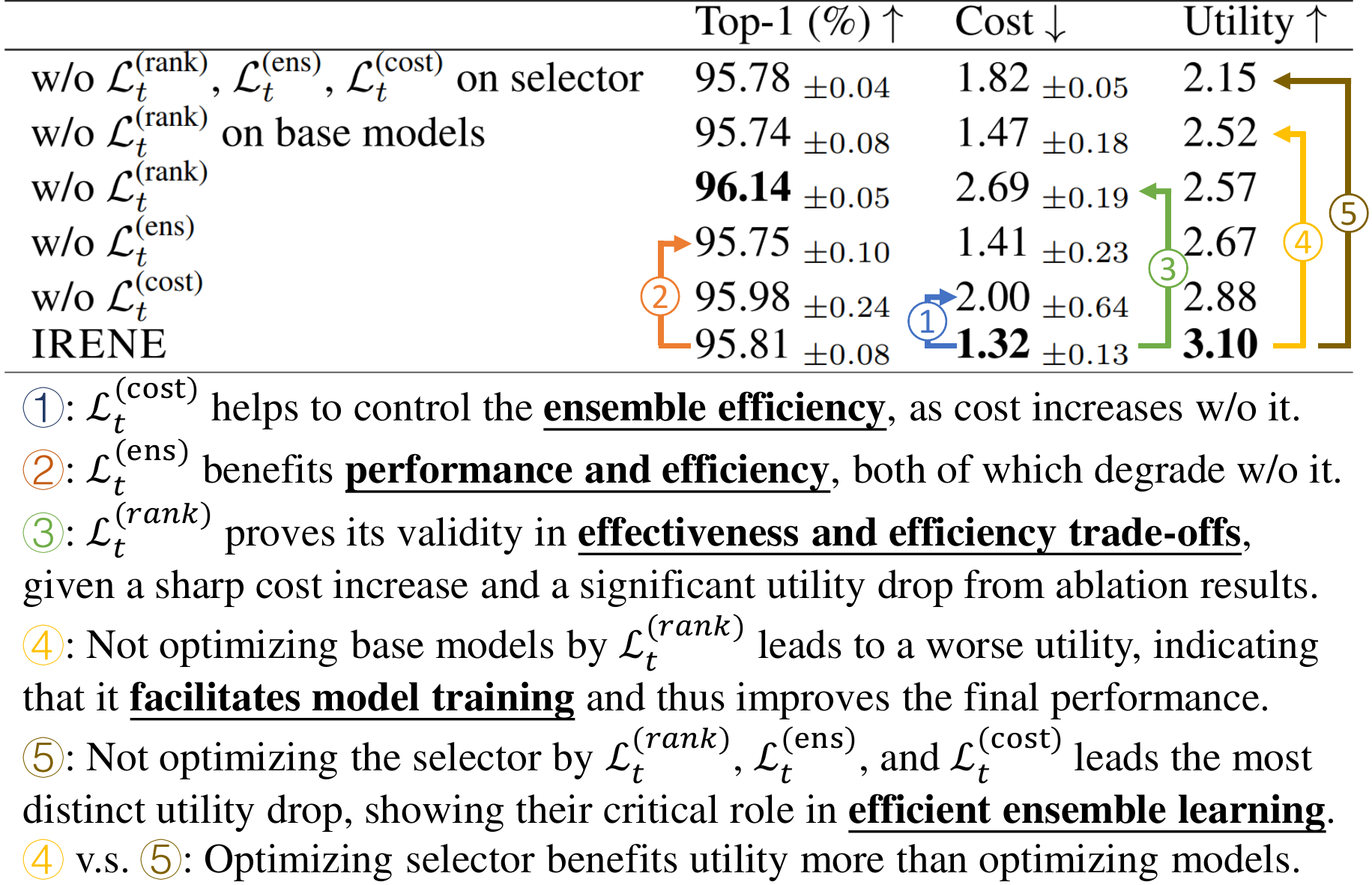} 
\caption{
Ablation study of the optimization objectives with ResNet-18 on CIFAR-10.
}
\label{fig: ablation}
\end{figure}

\subsubsection{Sensitivity analysis C: Ablation studies}

\noindent
We perform ablation studies to analyze the effects of our propose objectives in Eq.~\eqref{eq:total_loss}, except for the indispensable base model optimization objective  $\mathcal{L}^{\text{(base)}}_t$ in ensemble learning, on performance.
Furthermore, we want to verify the effectiveness of \method for (i) optimal halting over base models trained independently of the selector and (ii) sequential model training only, resp. 
Thus, we perform ablation experiments for the optimization of the selector and for the optimization of the base model.
To be specific, the selector is optimized by $\mathcal{L}^{\text{(rank)}}_t$, $\mathcal{L}^{\text{(ens)}}_t$, and $\mathcal{L}^{\text{(cost)}}_t$ together, while base models is optimized by $\mathcal{L}^{\text{(rank)}}_t$.
Figure~\ref{fig: ablation} presents the ablation results with detailed descriptions.
As shown in the figure, all the learning objectives play an indispensable role and the optimization of selector benefits the final performance more than that of base models.
Note that in all ablation settings, \method still achieves higher utility than all the baselines in Table~\ref{tab: cifar10}. 

\section{Conclusion}

In this paper, we focus on balancing the trade-off between ensemble effectiveness and efficiency, which is largely overlooked by existing ensemble approaches.
By modeling efficient ensemble inference as an optimal halting problem, we propose an \textit{effectiveness and efficiency-aware} selector network that is optimized jointly with base models via novel optimization objectives, to determine the halting strategies.
We demonstrate that \method can significantly reduce the inference cost while maintaining comparable performance to full ensembles, and beats existing computation-saving methods.
Optimal halting modeling also offers the possibility of solving problems such as sequential model selection with skipping to further boost computational savings, which we leave for further work.

\bibliography{aaai23}

\begin{thebibliography}{51}
\providecommand{\natexlab}[1]{#1}

\bibitem[{Abbasi et~al.(2020)Abbasi, Rajabi, Gagn{\'e}, and
  Bobba}]{abbasi2020toward}
Abbasi, M.; Rajabi, A.; Gagn{\'e}, C.; and Bobba, R.~B. 2020.
\newblock Toward adversarial robustness by diversity in an ensemble of
  specialized deep neural networks.
\newblock In \emph{Canadian Conference on Artificial Intelligence}, 1--14.
  Springer.

\bibitem[{Allen-Zhu and Li(2020)}]{allen2020towards}
Allen-Zhu, Z.; and Li, Y. 2020.
\newblock Towards understanding ensemble, knowledge distillation and
  self-distillation in deep learning.
\newblock \emph{arXiv preprint arXiv:2012.09816}.

\bibitem[{Bengio et~al.(2015)Bengio, Bacon, Pineau, and
  Precup}]{bengio2015conditional}
Bengio, E.; Bacon, P.-L.; Pineau, J.; and Precup, D. 2015.
\newblock Conditional computation in neural networks for faster models.
\newblock \emph{arXiv preprint arXiv:1511.06297}.

\bibitem[{Bengio, L{\'e}onard, and Courville(2013)}]{bengio2013estimating}
Bengio, Y.; L{\'e}onard, N.; and Courville, A. 2013.
\newblock Estimating or propagating gradients through stochastic neurons for
  conditional computation.
\newblock \emph{arXiv preprint arXiv:1308.3432}.

\bibitem[{Chen and Guestrin(2016)}]{chen2016xgboost}
Chen, T.; and Guestrin, C. 2016.
\newblock Xgboost: A scalable tree boosting system.
\newblock In \emph{Proceedings of the 22nd acm sigkdd international conference
  on knowledge discovery and data mining}, 785--794.

\bibitem[{Cho et~al.(2022)Cho, Kim, Jung, and Kweon}]{cho2022mcdal}
Cho, J.~W.; Kim, D.-J.; Jung, Y.; and Kweon, I.~S. 2022.
\newblock Mcdal: Maximum classifier discrepancy for active learning.
\newblock \emph{IEEE transactions on neural networks and learning systems}.

\bibitem[{Deb(2014)}]{deb2014multi}
Deb, K. 2014.
\newblock Multi-objective optimization.
\newblock In \emph{Search methodologies}, 403--449. Springer.

\bibitem[{DeVries and Taylor(2017)}]{devries2017improved}
DeVries, T.; and Taylor, G.~W. 2017.
\newblock Improved regularization of convolutional neural networks with cutout.
\newblock \emph{arXiv preprint arXiv:1708.04552}.

\bibitem[{Esser et~al.(2019)Esser, McKinstry, Bablani, Appuswamy, and
  Modha}]{esser2019learned}
Esser, S.~K.; McKinstry, J.~L.; Bablani, D.; Appuswamy, R.; and Modha, D.~S.
  2019.
\newblock Learned step size quantization.
\newblock \emph{arXiv preprint arXiv:1902.08153}.

\bibitem[{Figurnov et~al.(2017)Figurnov, Collins, Zhu, Zhang, Huang, Vetrov,
  and Salakhutdinov}]{figurnov2017spatially}
Figurnov, M.; Collins, M.~D.; Zhu, Y.; Zhang, L.; Huang, J.; Vetrov, D.; and
  Salakhutdinov, R. 2017.
\newblock Spatially adaptive computation time for residual networks.
\newblock In \emph{Proceedings of the IEEE conference on computer vision and
  pattern recognition}, 1039--1048.

\bibitem[{Frankle and Carbin(2018)}]{frankle2018lottery}
Frankle, J.; and Carbin, M. 2018.
\newblock The lottery ticket hypothesis: Finding sparse, trainable neural
  networks.
\newblock \emph{arXiv preprint arXiv:1803.03635}.

\bibitem[{Freund(1995)}]{freund1995boosting}
Freund, Y. 1995.
\newblock Boosting a weak learning algorithm by majority.
\newblock \emph{Information and computation}, 121(2): 256--285.

\bibitem[{Garipov et~al.(2018)Garipov, Izmailov, Podoprikhin, Vetrov, and
  Wilson}]{garipov2018loss}
Garipov, T.; Izmailov, P.; Podoprikhin, D.; Vetrov, D.; and Wilson, A.~G. 2018.
\newblock Loss surfaces, mode connectivity, and fast ensembling of dnns.
\newblock In \emph{Proceedings of the 32nd International Conference on Neural
  Information Processing Systems}, 8803--8812.

\bibitem[{Gontijo-Lopes, Dauphin, and
  Cubuk(2021)}]{gontijolopes2021representation}
Gontijo-Lopes, R.; Dauphin, Y.; and Cubuk, E.~D. 2021.
\newblock No One Representation to Rule Them All: Overlapping Features of
  Training Methods.
\newblock arXiv:2110.12899.

\bibitem[{Han, Mao, and Dally(2015)}]{han2015deep}
Han, S.; Mao, H.; and Dally, W.~J. 2015.
\newblock Deep compression: Compressing deep neural networks with pruning,
  trained quantization and huffman coding.
\newblock \emph{arXiv preprint arXiv:1510.00149}.

\bibitem[{Han et~al.(2015)Han, Pool, Tran, and Dally}]{han2015learning}
Han, S.; Pool, J.; Tran, J.; and Dally, W. 2015.
\newblock Learning both weights and connections for efficient neural network.
\newblock \emph{Advances in neural information processing systems}, 28.

\bibitem[{Han et~al.(2021)Han, Huang, Song, Yang, Wang, and
  Wang}]{han2021dynamic}
Han, Y.; Huang, G.; Song, S.; Yang, L.; Wang, H.; and Wang, Y. 2021.
\newblock Dynamic neural networks: A survey.
\newblock \emph{IEEE Transactions on Pattern Analysis and Machine
  Intelligence}.

\bibitem[{He et~al.(2016)He, Zhang, Ren, and Sun}]{he2016deep}
He, K.; Zhang, X.; Ren, S.; and Sun, J. 2016.
\newblock Deep residual learning for image recognition.
\newblock In \emph{Proceedings of the IEEE conference on computer vision and
  pattern recognition}, 770--778.

\bibitem[{Hochreiter and Schmidhuber(1997)}]{hochreiter1997long}
Hochreiter, S.; and Schmidhuber, J. 1997.
\newblock Long short-term memory.
\newblock \emph{Neural computation}, 9(8): 1735--1780.

\bibitem[{Hu et~al.(2020)Hu, Chen, Wang, and Wang}]{hu2020triple}
Hu, T.-K.; Chen, T.; Wang, H.; and Wang, Z. 2020.
\newblock Triple wins: Boosting accuracy, robustness and efficiency together by
  enabling input-adaptive inference.
\newblock \emph{arXiv preprint arXiv:2002.10025}.

\bibitem[{Huang et~al.(2017{\natexlab{a}})Huang, Chen, Li, Wu, Van Der~Maaten,
  and Weinberger}]{huang2017multi}
Huang, G.; Chen, D.; Li, T.; Wu, F.; Van Der~Maaten, L.; and Weinberger, K.~Q.
  2017{\natexlab{a}}.
\newblock Multi-scale dense networks for resource efficient image
  classification.
\newblock \emph{arXiv preprint arXiv:1703.09844}.

\bibitem[{Huang et~al.(2017{\natexlab{b}})Huang, Li, Pleiss, Liu, Hopcroft, and
  Weinberger}]{huang2017snapshot}
Huang, G.; Li, Y.; Pleiss, G.; Liu, Z.; Hopcroft, J.~E.; and Weinberger, K.~Q.
  2017{\natexlab{b}}.
\newblock Snapshot ensembles: Train 1, get m for free.
\newblock \emph{arXiv preprint arXiv:1704.00109}.

\bibitem[{Inoue(2019)}]{inoue2019adaptive}
Inoue, H. 2019.
\newblock Adaptive ensemble prediction for deep neural networks based on
  confidence level.
\newblock In \emph{The 22nd International Conference on Artificial Intelligence
  and Statistics}, 1284--1293. PMLR.

\bibitem[{Jacob et~al.(2018)Jacob, Kligys, Chen, Zhu, Tang, Howard, Adam, and
  Kalenichenko}]{jacob2018quantization}
Jacob, B.; Kligys, S.; Chen, B.; Zhu, M.; Tang, M.; Howard, A.; Adam, H.; and
  Kalenichenko, D. 2018.
\newblock Quantization and training of neural networks for efficient
  integer-arithmetic-only inference.
\newblock In \emph{Proceedings of the IEEE conference on computer vision and
  pattern recognition}, 2704--2713.

\bibitem[{Jang, Gu, and Poole(2016)}]{jang2016categorical}
Jang, E.; Gu, S.; and Poole, B. 2016.
\newblock Categorical reparameterization with gumbel-softmax.
\newblock \emph{arXiv preprint arXiv:1611.01144}.

\bibitem[{Ke et~al.(2017)Ke, Meng, Finley, Wang, Chen, Ma, Ye, and
  Liu}]{ke2017lightgbm}
Ke, G.; Meng, Q.; Finley, T.; Wang, T.; Chen, W.; Ma, W.; Ye, Q.; and Liu,
  T.-Y. 2017.
\newblock Lightgbm: A highly efficient gradient boosting decision tree.
\newblock \emph{Advances in neural information processing systems}, 30:
  3146--3154.

\bibitem[{Krizhevsky et~al.(2009)}]{krizhevsky2009learning}
Krizhevsky, A.; et~al. 2009.
\newblock Learning multiple layers of features from tiny images.

\bibitem[{Lakshminarayanan, Pritzel, and
  Blundell(2016)}]{lakshminarayanan2016simple}
Lakshminarayanan, B.; Pritzel, A.; and Blundell, C. 2016.
\newblock Simple and scalable predictive uncertainty estimation using deep
  ensembles.
\newblock \emph{arXiv preprint arXiv:1612.01474}.

\bibitem[{Lee et~al.(2015)Lee, Purushwalkam, Cogswell, Crandall, and
  Batra}]{lee2015m}
Lee, S.; Purushwalkam, S.; Cogswell, M.; Crandall, D.; and Batra, D. 2015.
\newblock Why M heads are better than one: Training a diverse ensemble of deep
  networks.
\newblock \emph{arXiv preprint arXiv:1511.06314}.

\bibitem[{Liang et~al.(2021)Liang, Glossner, Wang, Shi, and
  Zhang}]{liang2021pruning}
Liang, T.; Glossner, J.; Wang, L.; Shi, S.; and Zhang, X. 2021.
\newblock Pruning and quantization for deep neural network acceleration: A
  survey.
\newblock \emph{Neurocomputing}, 461: 370--403.

\bibitem[{Liu and Deng(2018)}]{liu2018dynamic}
Liu, L.; and Deng, J. 2018.
\newblock Dynamic deep neural networks: Optimizing accuracy-efficiency
  trade-offs by selective execution.
\newblock In \emph{Proceedings of the AAAI Conference on Artificial
  Intelligence}, volume~32.

\bibitem[{Malinin, Mlodozeniec, and Gales(2019)}]{malinin2019ensemble}
Malinin, A.; Mlodozeniec, B.; and Gales, M. 2019.
\newblock Ensemble distribution distillation.
\newblock \emph{arXiv preprint arXiv:1905.00076}.

\bibitem[{Mallya, Davis, and Lazebnik(2018)}]{mallya2018piggyback}
Mallya, A.; Davis, D.; and Lazebnik, S. 2018.
\newblock Piggyback: Adapting a single network to multiple tasks by learning to
  mask weights.
\newblock In \emph{Proceedings of the European Conference on Computer Vision
  (ECCV)}, 67--82.

\bibitem[{Rame and Cord(2021)}]{rame2021dice}
Rame, A.; and Cord, M. 2021.
\newblock Dice: Diversity in deep ensembles via conditional redundancy
  adversarial estimation.
\newblock \emph{arXiv preprint arXiv:2101.05544}.

\bibitem[{Sandler et~al.(2018)Sandler, Howard, Zhu, Zhmoginov, and
  Chen}]{sandler2018mobilenetv2}
Sandler, M.; Howard, A.; Zhu, M.; Zhmoginov, A.; and Chen, L.-C. 2018.
\newblock Mobilenetv2: Inverted residuals and linear bottlenecks.
\newblock In \emph{Proceedings of the IEEE conference on computer vision and
  pattern recognition}, 4510--4520.

\bibitem[{Sanh, Wolf, and Rush(2020)}]{sanh2020movement}
Sanh, V.; Wolf, T.; and Rush, A. 2020.
\newblock Movement pruning: Adaptive sparsity by fine-tuning.
\newblock \emph{Advances in Neural Information Processing Systems}, 33:
  20378--20389.

\bibitem[{Schuster and Paliwal(1997)}]{schuster1997bidirectional}
Schuster, M.; and Paliwal, K.~K. 1997.
\newblock Bidirectional recurrent neural networks.
\newblock \emph{IEEE transactions on Signal Processing}, 45(11): 2673--2681.

\bibitem[{Shazeer et~al.(2017)Shazeer, Mirhoseini, Maziarz, Davis, Le, Hinton,
  and Dean}]{shazeer2017outrageously}
Shazeer, N.; Mirhoseini, A.; Maziarz, K.; Davis, A.; Le, Q.; Hinton, G.; and
  Dean, J. 2017.
\newblock Outrageously large neural networks: The sparsely-gated
  mixture-of-experts layer.
\newblock \emph{arXiv preprint arXiv:1701.06538}.

\bibitem[{Shen et~al.(2020)Shen, Wang, Xu, Fu, Wang, and
  Lin}]{shen2020fractional}
Shen, J.; Wang, Y.; Xu, P.; Fu, Y.; Wang, Z.; and Lin, Y. 2020.
\newblock Fractional skipping: Towards finer-grained dynamic cnn inference.
\newblock In \emph{Proceedings of the AAAI Conference on Artificial
  Intelligence}, volume~34, 5700--5708.

\bibitem[{Tan et~al.(2019)Tan, Chen, Pang, Vasudevan, Sandler, Howard, and
  Le}]{tan2019mnasnet}
Tan, M.; Chen, B.; Pang, R.; Vasudevan, V.; Sandler, M.; Howard, A.; and Le,
  Q.~V. 2019.
\newblock Mnasnet: Platform-aware neural architecture search for mobile.
\newblock In \emph{Proceedings of the IEEE/CVF Conference on Computer Vision
  and Pattern Recognition}, 2820--2828.

\bibitem[{Teerapittayanon, McDanel, and
  Kung(2016)}]{teerapittayanon2016branchynet}
Teerapittayanon, S.; McDanel, B.; and Kung, H.-T. 2016.
\newblock Branchynet: Fast inference via early exiting from deep neural
  networks.
\newblock In \emph{2016 23rd International Conference on Pattern Recognition
  (ICPR)}, 2464--2469. IEEE.

\bibitem[{Thulasidasan et~al.(2019)Thulasidasan, Chennupati, Bilmes,
  Bhattacharya, and Michalak}]{thulasidasan2019mixup}
Thulasidasan, S.; Chennupati, G.; Bilmes, J.~A.; Bhattacharya, T.; and
  Michalak, S. 2019.
\newblock On mixup training: Improved calibration and predictive uncertainty
  for deep neural networks.
\newblock \emph{Advances in Neural Information Processing Systems}, 32.

\bibitem[{Wang et~al.(2021)Wang, Kondratyuk, Christiansen, Kitani,
  Movshovitz-Attias, and Eban}]{wang2021wisdom}
Wang, X.; Kondratyuk, D.; Christiansen, E.; Kitani, K.~M.; Movshovitz-Attias,
  Y.; and Eban, E. 2021.
\newblock Wisdom of Committees: An Overlooked Approach To Faster and More
  Accurate Models.
\newblock In \emph{International Conference on Learning Representations}.

\bibitem[{Wang et~al.(2018)Wang, Xu, Xu, and Tao}]{wang2018adversarial}
Wang, Y.; Xu, C.; Xu, C.; and Tao, D. 2018.
\newblock Adversarial learning of portable student networks.
\newblock In \emph{Proceedings of the AAAI Conference on Artificial
  Intelligence}, volume~32.

\bibitem[{Wen, Tran, and Ba(2020)}]{wen2020batchensemble}
Wen, Y.; Tran, D.; and Ba, J. 2020.
\newblock Batchensemble: an alternative approach to efficient ensemble and
  lifelong learning.
\newblock \emph{arXiv preprint arXiv:2002.06715}.

\bibitem[{Yang et~al.(2022)Yang, Ren, Luo, Liu, Liu, Bian, Zhang, and
  Li}]{yang2022towards}
Yang, Z.; Ren, K.; Luo, X.; Liu, M.; Liu, W.; Bian, J.; Zhang, W.; and Li, D.
  2022.
\newblock Towards Applicable Reinforcement Learning: Improving the
  Generalization and Sample Efficiency with Policy Ensemble.
\newblock \emph{arXiv preprint arXiv:2205.09284}.

\bibitem[{Zhang, Liu, and Yan(2020)}]{zhang2020diversified}
Zhang, S.; Liu, M.; and Yan, J. 2020.
\newblock The diversified ensemble neural network.
\newblock \emph{Advances in Neural Information Processing Systems}, 33.

\bibitem[{Zhang et~al.(2018)Zhang, Zhou, Lin, and Sun}]{zhang2018shufflenet}
Zhang, X.; Zhou, X.; Lin, M.; and Sun, J. 2018.
\newblock Shufflenet: An extremely efficient convolutional neural network for
  mobile devices.
\newblock In \emph{Proceedings of the IEEE conference on computer vision and
  pattern recognition}, 6848--6856.

\bibitem[{Zhou, Wang, and Bilmes(2018)}]{zhou2018diverse}
Zhou, T.; Wang, S.; and Bilmes, J.~A. 2018.
\newblock Diverse ensemble evolution: Curriculum data-model marriage.
\newblock In \emph{Proceedings of the 32nd International Conference on Neural
  Information Processing Systems}, 5909--5920.

\bibitem[{Zhou et~al.(2020)Zhou, Xu, Ge, McAuley, Xu, and Wei}]{zhou2020bert}
Zhou, W.; Xu, C.; Ge, T.; McAuley, J.; Xu, K.; and Wei, F. 2020.
\newblock Bert loses patience: Fast and robust inference with early exit.
\newblock \emph{Advances in Neural Information Processing Systems}, 33:
  18330--18341.

\bibitem[{Zhou, Wu, and Tang(2002)}]{zhou2002ensembling}
Zhou, Z.-H.; Wu, J.; and Tang, W. 2002.
\newblock Ensembling neural networks: many could be better than all.
\newblock \emph{Artificial intelligence}, 137(1-2): 239--263.

\end{thebibliography}

\clearpage

\section{A Implementation Details}

In this section, we describe the implementation details including network architectures and hyper-parameter tuning.
We will release the source code of all the experiments upon paper acceptance.

\subsection{A.1 Our proposed method \method}

For \method, we present its implementation in terms of the two main components it contains, namely the selector and base models.
And the learning curve is shown on Figure~\ref{fig: learning_curve} with analysis in Appendix D.

\subsubsection{A.1.1 Selector network implementation}

\quad

\noindent
\textbf{Architecture.}
The selector network uses the standard LSTM architecture~\cite{hochreiter1997long}, that takes $t$-th model outcomes and the hidden state vector $d_{t-1}$ of the last timestep to output $h_t$ at each timestep $t$.
The initial state $d_{0}$ is initialized as all zeros.
Note that the selector is lightweight, e.g., with 0.000004 of the number of parameters of ResNet-18.

\noindent
\textbf{Learning.}
We use Adam optimizer whose learning rate and weight decay are hyper-parameters to be tuned.
As the selector is jointly trained with image backbones which generally use a specific learning rate scheduling, we adopt a standard scheduler that decreases the learning rate at steps that schedulers of backbone set.
Additionally, we will re-initialize the optimizer and scheduler at each timestep.
And we use a different multiplicative factor, as a hyper-parameter, of the learning rate decay.

\noindent
\textbf{Hyper-parameters.}
We summarize the hyper-parameters with their search range in Table~\ref{table: hyper-parameters}.
For all datasets, we run the hyper-parameter tuning with 24 search steps. 

\begin{table}[]
\centering
\caption{
Hyper-parameters we tune for the selector network. 
We search learning rates and loss weights using log uniform.
}
\begin{tabular}{|l|l|l|l|}
\hline
Hyper-parameter & Search Range  \\ \hline
learning rate  & [1e-5,1e-1]  \\ \hline
learning rate decay  & [0.1, 0.2, 0.5, 0.8]  \\ \hline
loss weight $\omega_1$ & [1e-4,0.1]  \\ \hline
loss weight $\omega_2$ & [1e-5,0.1]  \\ \hline
loss weight $\omega_3$ & [1e-4,0.1]  \\ \hline
\end{tabular}
\label{table: hyper-parameters}
\end{table}

\subsubsection{A.1.2 Base model implementation}

\quad

\noindent
\textbf{Architecture.}
We use ResNet-32 and ResNet-18 as backbones, whose predictions are fed into selector to predict the optimal halting probability.
However, predictions from a single model can be misleading, such as they are prone to overconfidence~\cite{thulasidasan2019mixup}.
While \citet{cho2022mcdal} shows that the discrepancy of linear classifiers on top of a shared feature extractor can help indicate the \textit{prediction uncertainty}.
Following their design, we add one additional classification head (1-layer perceptron) and realize the discrepancy of the two classifiers by maximizing the L1 distance between their output predictions, which served as a regularization term.
Based on the predictions of the original and additional classification heads, we compute their KL-divergence to illustrate the prediction uncertainty of this base model, and feed the results into the selector $g_{\phi}$.
Then, the selector will decide the halting probability based on the current inference situations.
Note that the computational cost of adding one classification head is negligible.
For example, adding one classifier head to ResNet-18~\cite{he2016deep} would only increase 0.0002 of the original model parameters.

\noindent
\textbf{Learning and hyper-parameter.}
On the CIFAR datasets, we adopt the standard parameters applied for these commonly used backbones~\cite{devries2017improved}.
Specifically, we use SGD as the optimizer, with a weight decay of 5e-4 and a Nesterov momentum of 0.9.
The initial learning rate is 0.1, which is divided by 5 for the 60-th, 120-th, and 160-th epochs.
The total training epochs are 200, with a batch size of 128.
By adopting the standard settings, the base models do not require any other hyper-parameter tuning except for the loss weight $\omega_3$ of $\mathcal{L}^{\text{(rank)}}_k$ listed in Table~\ref{table: hyper-parameters} in Appendix A.1.1.

\subsection{A.2 Baselines}

We here introduce the implementation of baselines.
Aligning the experimental setups, we run all baselines, except for Div$^{2}$ whose results are taken from their original paper.
We list all the baseline implementation we use as follows.

\begin{itemize}
    \item Average ensemble: we train different initialized models and aggregate their predictions averagely.
    \item Snapshot ensemble and FGE: we use the implementation from torchensemble\footnote{\url{https://github.com/TorchEnsemble-Community/Ensemble-Pytorch#id14}}.
    \item MoE: we implemente a general MoE framework, where the gating network is realized as $G(x) = \text{Softmax}(E(x) \cdot \mathbf{W_g})$ and $E(\cdot)$ is a fixed feature extractor (corresponding to the backbone) from torchvision\footnote{\url{https://pytorch.org/docs/stable/model_zoo.html#module-torch.utils.model_zoo}} and $\mathbf{W_g}$ is a trainable weight matrix. 
    \item Sparse-MoE: we use a third-party implementation\footnote{\url{https://github.com/lucidrains/mixture-of-experts}}.
    \item WoC: we take models trained for average ensemble, on top of which to determine whether to stop inference at each timestep using their confidence threschold strategy~\cite{wang2021wisdom}.
    In the original paper, they conduct a grid search for predefined thresholds.
    Accordingly, we perform a parameter search by setting the search value to start at 0.0 and end at 1.0 with an interval of 0.01.

\end{itemize}

\section{B Training Algorithm}

Algorithm~\ref{our-algorithm} is our sequential training process.

\begin{algorithm}
 \caption{Training algorithm}
 \label{our-algorithm}
  \KwIn{Models $\{f_{\theta_t}\}_{t=1}^T$, selector $g_{\phi}$, training epochs $O$}
  \KwOut{}

  \For{$t=1$ to $T$}{
    \For{$i=1$ to $O$}{
    $f_{\theta_t}$ generates prediction $\hat{y}_t$
    
    $g_{\phi}$ takes $\hat{y}_t$ and outputs $h_t$
    
    Compute $p_t$ based on $h_t$ via Eq.~\eqref{eq:p_t}
    
    Calculate objectives via Eq.~\eqref{eq:total_loss}
    
    Update $\phi$ and $\theta_t$ via gradient descent}

  }
\end{algorithm}

\begin{figure*}[h]
\centering
\vspace{-0.0em}
\includegraphics[width=0.75\textwidth]{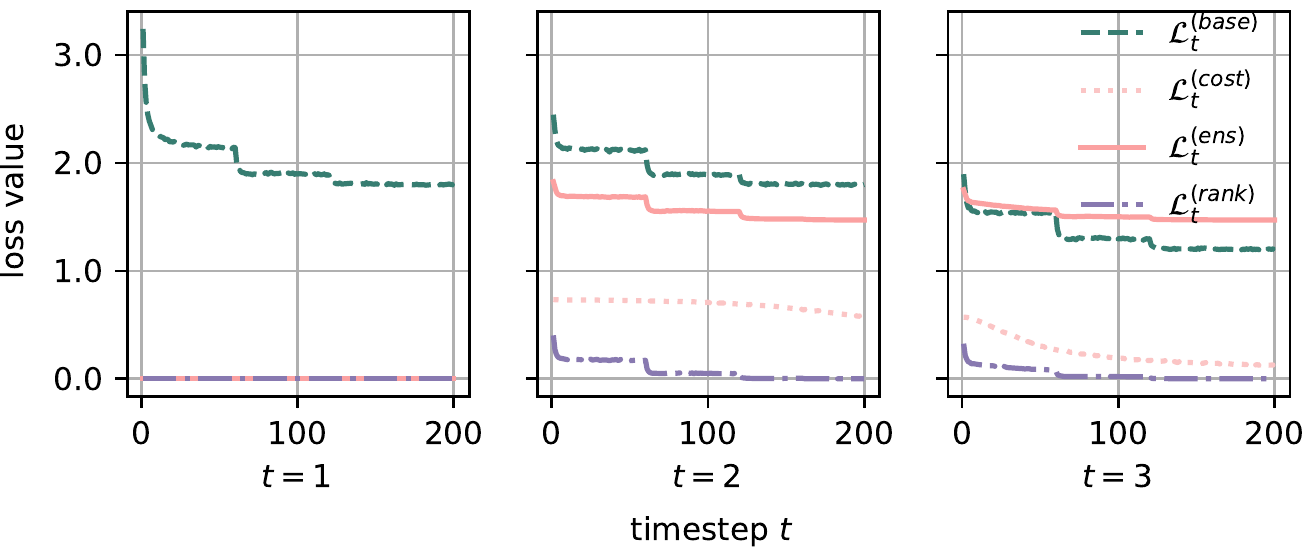} 
\caption{
Learning curves of our optimization process.
}
\label{fig: learning_curve}
\end{figure*}

\section{C Evaluation Metrics}

Following \citet{tan2019mnasnet}, we use a customized weighted product method to calculate the trade-off between ensemble performance and efficiency as
\begin{equation}\label{eq:utility}
\begin{aligned}
  \text{utility} = \left[\dfrac{V(\cdot)}{v} \right]^{\tau_{v}}  \div \left[ \dfrac{C(\cdot)}{c}\right]^{\tau_{c}},
\end{aligned}
\end{equation}
where $v$ and $c$ are the target performance and cost, and $\tau_v$ and $\tau_{c}$ are the weight factors.
For fair comparison, we keep the parameters the same in utility calculation for all the compared methods on the same dataset. 

Here we detail the calculation of utility metric and we start by describing the coefficient calculation.
For simplicity, $v$ and $c$ are set to 1.
For $\tau_v$ and $\tau_c$,
we follow \cite{tan2019mnasnet} to determine empirically by ensuring different Pareto optimal solutions~\cite{deb2014multi} have the same metric value.
Based on the observation that average ensemble generally yields the best performance among baselines using the same computational cost (Table~\ref{tab: cifar10} and ~\ref{tab: cifar100}), we can suppose that average ensemble can obtain the Pareto optimal values when using $T$ units of cost cost.
Naturally, the single model is the Pareto optimal solution when using one unit of cost, under the problem setting.
Therefore, we can calculate $\tau$ values based on the utility equality of the single model and the average ensemble (with $T = 3$).
In practice, $\tau_c$ is similar under different settings.
We set $\tau_c = 0.01$ based on empirical evidence and then calculate $\tau_v$.
Note that $\tau_v$ is different for the different combinations of datasets and backbones used.

To make the calculated utility values more obviously comparable, we further calculate the difference between the utility value of each method with that of the single model, and take the exponential form of the calculated differences making the differences clearer.

We also report top-1 accuracy \textcolor{black}{(i.e., the accuracy where true class matches with the most probable classes predicted by the ensemble)} and the corresponding average inference cost, i.e., average number of utilized models in ensemble.

\section{D Learning curve}

In this section, we present learning curves of the optimization objectives used by \method, to verify and help understand the learning situations of \method.
Curves of all timesteps are shown in Figure~\ref{fig: learning_curve}.
Note that when $t=1$, only $\mathcal{L}^{\text{(base)}}_t$ is optimized.

From the figure, the overall optimization is stable, although various objectives have been involved.
Additionally, we find that base model training benefits from its joint optimization with the selector by comparing the curves of $\mathcal{L}^{\text{(base)}}_t$ in different timesteps.
Specifically, when $t=1$, there is no difference with the training of a single model.
After that, base models are additionally optimized by $\mathcal{L}^{\text{(rank)}}_t$ which encourages base models to focus on samples assigned to them.
From the figure, the loss values for the later timesteps are lower than the previous one(s) and also converge to lower values.
It indicates the benefit of joint optimization for the base models and the selector, and the learning of our method is stable.

\end{document}